\pdfoutput=1

\documentclass[11pt]{article}

\usepackage{acl}

\usepackage{times}
\usepackage{latexsym}

\usepackage[T1]{fontenc}

\usepackage[utf8]{inputenc}

\usepackage{microtype}

\usepackage{inconsolata}

\usepackage{graphicx}
\usepackage{tabularx}
\usepackage{amsmath}
\usepackage{multirow}

\title{Using LLMs to identify features of personal and professional skills in an open-response situational judgment test}

\author{Cole Walsh, Rodica Ivan, Muhammad Zafar Iqbal, and Colleen Robb \\
  Acuity Insights Inc., Toronto, ON, Canada}

\begin{document}
\maketitle
\begin{abstract}
Academic programs are increasingly recognizing the importance of personal and professional skills and their critical role alongside technical expertise in preparing students for future success in diverse career paths. With this growing demand comes the need for scalable systems to measure, evaluate, and develop these skills. Situational Judgment Tests (SJTs) offer one potential avenue for measuring these skills in a standardized and reliable way, but open-response SJTs have traditionally relied on trained human raters for evaluation, presenting operational challenges to delivering SJTs at scale. Past attempts at developing NLP-based scoring systems for SJTs have fallen short due to issues with construct validity of these systems. In this article, we explore a novel approach to extracting construct-relevant features from SJT responses using large language models (LLMs). We use the Casper SJT to demonstrate the efficacy of this approach. This study sets the foundation for future developments in automated scoring for personal and professional skills.
\end{abstract}

\section{Background}

A longstanding challenge in academia is selecting qualified and professional candidates from a larger applicant pool for professional training programs. Decision makers have traditionally relied on measures of hard skills and cognitive ability to make these decisions~\cite{eva2009predictive}, often relying on grade point average (GPA) and standardized tests such as the Scholastic Aptitude Test (SAT), Graduate Record Exam (GRE), Medical College Admission Test (MCAT), and Graduate Management Admission Test (GMAT). Personal and professional skills such as communication, teamwork, problem-solving, and critical thinking, although recognized as predictive of future success in education and industry~\cite{heckman2012hard}, have been more difficult to measure for a number of reasons including lack of standardization and scalability~\cite{patterson2016effective}. Admissions committees have commonly used reference letters, personal essays, and interviews as a proxy for applicants’ personal and professional skills, but these processes do not meet the psychometric standards that we would expect from tools used in high-stakes decision-making~\cite{kuncel2014meta, patterson2016effective}. Additionally, as the adoption of generative AI spreads, there are increased concerns about the authenticity of reference letters and personal essays~\cite{chen2024can}, further exacerbating the need for valid and reliable tools to measure personal and professional skills.

Recognizing the limitations of other admissions tools (i.e., reference letters, personal essays)~\cite{patterson2016effective}, higher education programs have been increasingly turning to a more reliable and standardized tool, Situational Judgment Tests (SJTs), to assess applicants' personal and professional skills as part of their admissions process~\cite{webster2020situational, nadmilail2023broad}. Though they may be delivered in different formats, including fixed-response and open-response, SJTs generally involve simulated situations and questions designed to elicit how a respondent would likely react in the situation~\cite{lievens2013adjusting}. Fixed-response SJTs typically require respondents to select or rank possible actions based on their effectiveness in a given situation and show stronger relationships with measures of cognitive ability, rather than personal or professional skills~\cite{mcdaniel2007situational}. Open-response SJTs, on the other hand, are more conducive to measuring behavioral tendencies (i.e., how the respondent would likely react in the given situation) and tend to show stronger relationships with personal and professional skills relative to fixed-response SJTs~\cite{mcdaniel2007situational}.

Although open-response SJTs have proven effective in evaluating personal and professional skills in a standardized and reliable manner, there are challenges in executing these types of assessments at scale. Open-response SJTs are primarily scored by human raters who require extensive training to become proficient at evaluating responses~\cite{shipper2017pilot}. Additionally, performing this level of scoring at scale requires many trained human raters operating in parallel, which presents further operational barriers. These challenges are not unique to SJTs; developers of other open-response assessments have faced similar obstacles and overcome them with automated scoring systems such as Natural Language Processing (NLP) algorithms~\cite{valenti2003overview}. NLP-based scoring systems of this kind have been shown to achieve strong psychometric results in writing and language proficiency tests~\cite{chodorow2004beyond, ramineni2012evaluation,  cardwell2022duolingo}, as well as short-answer tasks~\cite{leacock2003c}.

While there is a growing literature on NLP-based scoring systems for open-response assessments, few studies have investigated their efficacy specifically for SJTs~\cite{bulut2022leveraging, walsh2022we}. One issue is that insights from other automated scoring systems may not be immediately transferable to SJTs given the difference in the measured construct: while other open-response assessments may focus on language proficiency or content-mastery, SJTs measure personal and professional skills (e.g., teamwork, problem solving, critical thinking)~\cite{lievens2016situational}. These differences in the measured construct influence the kinds of features used as inputs to the scoring system. In particular, NLP-based scoring systems for writing and language proficiency typically use features related to coherence, grammar, and organization~\cite{chodorow2004beyond, ramineni2012evaluation,  cardwell2022duolingo}, features which have no direct link with most constructs assessed by SJTs. Any valid NLP-based scoring system should exhibit construct relevance through the features used as inputs to said system~\cite{mccaffrey2022best}, making existing approaches to NLP-based scoring largely inapplicable to SJTs. Additionally, because open-response SJTs allow respondents to describe actions that they would take or have taken in the past, these assessments are designed to allow for complexity and response diversity, and thus there is generally no single correct answer~\cite{dore2017casper}. This characteristic of SJTs makes scoring responses based on "correctness" or similarity with other responses impractical as well.

\section{Aims}
In this study, we investigate the feasibility of identifying and extracting construct-relevant features from SJT responses. We build on the work of Iqbal \textit{et al.}\cite{iqbal2025evaluating} who used a mixed-methods approach to identify nine construct-relevant features that influenced raters' evaluations of an open-response SJT. We probe whether and to what extent we can identify these features in SJT responses automatically using NLP-based approaches. Recognizing the complex and nuanced nature of these features, we decided to use Large Language Models (LLMs) for this task. Recent studies have demonstrated strong performance of LLMs for essay scoring~\cite{lee2024applying} even in domains like divergent thinking~\cite{organisciak2023beyond}, which, similar to SJTs, have been notoriously difficult to automatically score because of the complex nature of the construct. This work sets the stage for future endeavors to build an automated scoring system for open-response SJTs and similar assessment types.

\section{Sample}

We used data from the Casper SJT in this study. Casper is an open-response SJT that purports to measure respondents' personal and professional skills along the following competencies: collaboration, communication, empathy, ethics, fairness, motivation, problem solving, resilience, and self-awareness~\cite{dore2017casper, saxena2024incorporating}. Casper presents respondents with a series of hypothetical scenarios that include either a text-based or video-based prompt. Text-based prompts include a short description of a situation while video-based prompts include trained actors enacting a scripted scenario. The respondents are then asked questions related to the prompt and given a fixed amount of time to respond. The data we used in this study included responses to both types of scenarios: video-based and word-based. An example of a situation depicted in a video-prompt scenario is given below:
\begin{quote}
    Chris and Jane are sitting together in a small meeting room. Their manager, Gary, enters to deliver a few brief comments before retreating to an adjoining work space. Chris gets up to approach Gary when he notices Gary focused on his phone instead of work. Jane tells Chris that she sees Gary on his phone very often and that overall he does not do a lot of work. Chris says it does not seem fair for someone like Gary, who is senior to them in the company, to do less work and be paid a lot more. In addition, Gary takes their hard work to present as his own, taking credit for their efforts.
\end{quote}
Respondents were instructed that they were a coworker of Chris and Jane in this scenario and asked the following three questions:
\begin{enumerate}
    \item How would you handle this situation with Gary, your manager? Explain your response.
    \item Imagine that Gary was completing his work in a timely manner outside of normal hours, but still behaving inappropriately while in the office. Would this change your opinion? Why or why not?
    \item Describe some serious issues that can occur when supervisors are not present for their team.
\end{enumerate}

In addition to the different types of scenario prompts, Casper also includes two distinct response formats: respondents are either required to type their responses to the associated questions within the allocated time or record a video of themselves responding to each question. To simplify this study, we only examined scenarios with the typed-response format as analyzing video responses would have required either transcribing the responses or passing the video media itself to a multimodal AI model, fundamentally altering the procedure employed here. We leave the analysis of video responses to a future study.

Casper is a completely human-rated assessment where a respondent's responses to a scenario are rated together holistically on a 1-9 Likert scale. Trained human raters are provided with scoring guidelines which help them contextualize Casper competencies for each scenario and determine how effectively the responses addressed the questions asked. Additionally, Casper is norm-referenced, which means that raters are also instructed to score each response relative to the other responses they are reviewing for the same scenario. Raters do not, however, use an analytical rubric when rating in order to encourage diverse perspectives and interpretation during the rating process. This rating approach allows for responses exhibiting different characteristics to still receive high scores as the context and reasoning provided by the respondent are also taken into account.

The diversity in scenarios and responses makes Casper ideally suited to study underlying features of personal and professional skills in SJT responses. Previously, Iqbal \textit{et al.} identified nine construct-relevant features that influenced Casper ratings. For the purposes of the current study, we selected seven of these features to investigate the applicability of LLMs for feature extraction, omitting two features related to competencies associated with specific scenarios. Given that different Casper competencies are probed in each scenario, we omitted these two features from our investigation as their analysis would have required prompting with more scenario and competency-specific information, which would have extended the complexity of this pilot study. 

Table~\ref{tab:features} shows the seven features we selected for this study. Iqbal \textit{et al.} previously analyzed 27 responses from each of three Casper typed-response scenarios, ensuring a uniform distribution of responses at each scoring level (i.e., three responses for each 1-9 score assigned by human raters). Two researchers independently classified the construct-relevant features present in all responses using the levels noted in Table~\ref{tab:features}.

\begin{table*}[ht]
    \centering
    \begin{tabularx}{\textwidth}{lXXc}
        \hline
        \textbf{Key} & \textbf{Description} & \textbf{Levels} & \textbf{Cohen's $\kappa$ (Humans)} \\
        \hline
        INT & Grasps and addresses the complex social and emotional dynamics present in the ethical dilemma. & 
        \begin{itemize}
            \item Limited Interpretation
            \item Adequate Interpretation
            \item Excellent Interpretation
        \end{itemize} & $0.700$\\
        \hline
        LACKINF & States that they do not have enough information to make a decision. & True/False & $0.640$ \\
        \hline
        JUST &  Justifies the course of action suggested. & 
        \begin{itemize}
            \item No Justification
            \item Superficial Justification
            \item Reasonable Justification
            \item Clear and Compelling Justification
        \end{itemize} & $0.788$\\
        \hline
        VAGUE & Vague or unclear. & True/False & $0.356$ \\
        \hline
        PERSP & Considers the perspectives of the different parties involved in the dilemma. & \begin{itemize}
            \item Considers one perspective
            \item Briefly considers multiple perspectives
            \item Thoughtfully considers multiple perspectives
        \end{itemize} & $0.722$\\
        \hline
        DISRES & Fails to acknowledge or show sensitivity towards the legitimate concerns or feelings of one of the parties involved. & True/False & $0.647$ \\
        \hline
        CREAT & Provides insightful, novel, or creative arguments to address the question. & True/False & $0.510$ \\
        \hline
    \end{tabularx}
    \caption{Features identified by Iqbal \textit{et al.} as influencing Casper scores and used in the present study. Cohen's $\kappa$ is reported between two independent human raters' classifications of these features across 162 Casper responses.}
    \label{tab:features}
\end{table*}

For the present study, we doubled the size of the dataset used, re-using the dataset collected by Iqbal \textit{et al.} while adding 27 responses from each of three additional scenarios, which were again classified by the same two human raters as in the original study. Thus, the complete dataset in this study comprised 162 responses across six distinct Casper scenarios. We report agreement for the researchers' classifications of the features across all 162 responses in the last column of Table~\ref{tab:features} for each of the seven features. We used Cohen's $\kappa$ with quadratic weighting (after mapping features to a numeric scale) to measure agreement. In the case of binary features, the quadratic-weighted $\kappa$ is identical to an unweighted $\kappa$.

\section{Methods}

We used LLMs as classifiers to replicate the work of the human raters in identifying construct-relevant features in Casper responses. Previous studies of LLMs for essay scoring have identified performance gains when LLMs are allowed to specifically evaluate one aspect of writing at a time~\cite{lee2024applying}. We applied a similar principle here by prompting LLMs to evaluate only one feature at a time for a given response.

We conducted two separate analyses. In our first study we compared several LLMs based on how well their classifications aligned with those of human raters for each feature. We used five state-of-the-art LLMs, listed in Table~\ref{tab:llms}, including a mix of reasoning and non-reasoning, open and closed-source models.

\begin{table*}[ht]
    \centering
    \caption{LLMs explored in this study including their providers, whether they were reasoning models, and whether model weights were open or closed-source.}
    \label{tab:llms}
    \begin{tabular}{l l c c}
        \hline
        \textbf{Name}   & \textbf{Provider} & \textbf{Reasoning Model (Y/N)} & \textbf{Open/closed-source} \\ \hline
        GPT-4o-mini  & OpenAI & N             & Closed               \\
        DeepSeek-R1   & DeepSeek & Y             & Open             \\
        Lllama 4 Maverick  & Meta & N             & Open             \\
        o4-mini  & OpenAI & Y             & Closed               \\
        Claude Sonnet 4 & Anthropic & Y            & Closed             \\ \hline
    \end{tabular}
\end{table*}

For each LLM and feature, we generated classifications for all 162 Casper responses, then computed the $\kappa$ between model classifications and each human rater's classifications. We then averaged the results to obtain one average $\kappa$ for each LLM and feature. We used the same zero-shot prompt for all LLMs; we did not provide the LLMs with any examples within the prompts. Further, we wanted to compare how well each LLM performed using the same prompt without tailoring the prompt to work better with one LLM or another, so we provided only the necessary information to carry out the task within the prompt. Below is a minimal reproducible example of the system prompt we used:
\begin{quote}\itshape
    You are a helpful assistant that analyzes users' responses to an ethical dilemma.
    
    The user was given the following prompt: "\{context\}".\\
    They were asked to respond to the following questions related to this prompt: "\{questions\}".
    
    Your task is to analyze \{feature\_description\}.\\
    Return your response as a JSON object with the following keys:\\
    "decision": <\{feature\_levels\}>\\
    "reasoning": <Reasoning for decision>
\end{quote}
where curly braces denote variable inputs in the prompt. \{context\} is the text prompt (for text-based scenarios) or the summary of the video (for video-based scenarios), \{questions\} is an enumerated list of the questions asked, \{feature\_description\} is a description of the feature being evaluated (see Table~\ref{tab:features}), and \{feature\_levels\} are the possible values that the LLM may assign.

We required the LLMs to return a "reasoning" for their decisions. For example, using the scenario and questions described above with the PERSP feature, o4-mini as the LLM, and the following response:
\begin{enumerate}
    \item I’d invite Gary to a friendly chat, share what Jane and I feel, and listen to his pressures. Together we’d create small habits—visible credit tags and quick check-ins—so everyone’s effort shines and Gary can still balance his senior duties.
    \item If Gary finishes work later, I’d appreciate his flexibility, yet gently note that daytime presence and fair credit nurture team spirit. Leadership isn’t only outcomes; it’s modeling engagement and celebrating others while they’re present.
    \item When supervisors vanish, teammates can feel adrift: goals drift, growth pauses, mistakes linger, and trust erodes. Consistent guidance, recognition, and availability are anchors that keep performance, learning, and well-being afloat.
\end{enumerate}
we get the following output:
\begin{quote}\itshape
    {
    "decision":"Thoughtfully considers and empathizes with multiple perspectives",\\
    "reasoning":"The user’s responses address the concerns of the co-workers (feeling undervalued and needing recognition), invite Gary to share his pressures and perspective, and propose collaborative solutions that respect his senior role. They also weigh the team’s needs for presence, credit, and guidance, demonstrating balanced empathy for both the manager and the team."
    }
\end{quote}
We do not inspect "reasoning" outputs in this study, though such components can provide important validity evidence for LLM-based scoring systems~\cite{casabianca2025validity}, hence why we include this element in our prompting strategy.

In our second study, we investigated whether and to what extent we could improve LLM-human agreement by providing further details, including inclusion and exclusion criteria, for each level of a feature. Previous studies have indicated that this prompting strategy can effectively improve LLM performance in essay evaluation tasks~\cite{lee2024applying}. We worked exclusively with o4-mini for this analysis because it offered the best combination of throughput, cost, and performance that was ideally suited for this iterative pilot study.

\section{Results}

\subsection{Comparison of LLMs with zero-shot prompt}

\begin{figure*}[t]
  \includegraphics[width=\linewidth]{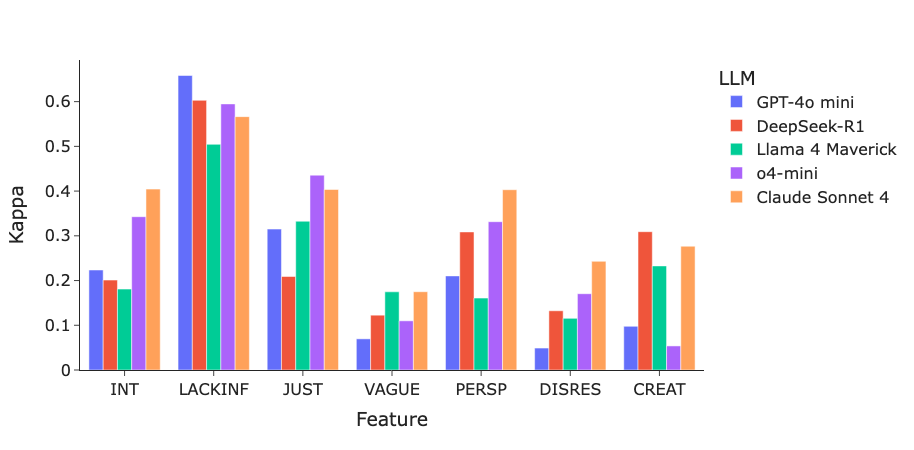}
  \caption {Average $\kappa$ with human raters using zero-shot prompt.}
  \label{fig:compare_llms}
\end{figure*}

Results are shown in Table~\ref{tab:results}, while Figure~\ref{fig:compare_llms} shows the average $\kappa$ agreement between each LLM and the two human raters on each feature using the zero-shot prompt. We find that Claude Sonnet 4 generally outperforms the other models, achieving the highest agreement with human raters on four out of seven features, while achieving the second highest agreement on two other features (JUST and CREAT). LACKINF was the lone feature where Claude Sonnet 4 did not rank among the top two LLMs, but even in this case the model achieved near human-level agreement ($\kappa_{\text{Claude Sonnet 4}} = 0.566$ compared to $\kappa_{\text{Humans}} = 0.640$).

\begin{table*}[ht]
    \centering
    \begin{tabular}{c|ccccc|c}
        \hline
        & \multicolumn{5}{c}{\textbf{Zero-shot}} & \textbf{Level Desc.} \\
        \textbf{Feature} & \textbf{GPT-4o mini} & \textbf{DeepSeek-R1} & \textbf{Llama 4 Mav.} & \textbf{o4-mini} & \textbf{Sonnet 4} & \textbf{o4-mini} \\ \hline
        INT & 0.224 & 0.201 & 0.181 & 0.343 & \textbf{0.404} & 0.434 \\
        LACKINF & \textbf{0.658} & 0.603 & 0.505 & 0.595 & 0.566 & - \\
        JUST & 0.315 & 0.209 & 0.333 & \textbf{0.436} & 0.404 & 0.479 \\
        VAGUE & 0.070 & 0.123 & \textbf{0.175} & 0.110 & \textbf{0.175} & 0.191 \\
        PERSP & 0.210 & 0.309 & 0.161 & 0.332 & \textbf{0.403} & 0.431 \\
        DISRES & 0.049 & 0.132 & 0.116 & 0.171 & \textbf{0.243} & 0.377 \\
        CREAT & 0.098 & \textbf{0.309} & 0.233 & 0.054 & 0.277 & 0.145 \\ \hline
    \end{tabular}
    \caption{Average Cohen's $\kappa$ agreement with human raters for each LLM on each feature using the zero-shot prompt. The last column shows the average $\kappa$ for o4-mini after modifying the prompts to include level descriptions for each feature. We did not explore the LACKINF feature in this second experiment because we achieved close to human-level agreement with the zero-shot prompt.}
    \label{tab:results}
\end{table*}

o4-mini provides similar results to Claude Sonnet 4 in most cases, but notably struggles to identify responses that "provide insightful, novel, or creative arguments to address the questions" (CREAT). DeepSeek-R1, on the other hand, excels at identifying responses that fit this definition. GPT-4o mini is generally outclassed by the other models, but does reach super-human agreement in identifying responses that "state that they do not have enough information to make a decision" (LACKINF). Among all features explored in this study, LACKINF is the most likely to be identifiable through the use of particular words or phrases. For example, the string "gather more information" appears in seven responses. One human rater marked four of these responses as exhibiting the LACKINF feature while the other human rater marked all seven responses as exhibiting this feature. GPT-4o mini, meanwhile, classified six out of the seven responses as exhibiting LACKINF. Features such as this one that may be identifiable through keyword or semantic relationships likely see smaller benefits from using LLMs and, especially, more advanced reasoning models.

For four out of seven features, the top performing LLM achieved $\kappa > 0.4$. However, none of the LLMs approached human-level agreement on any feature outside of LACKINF. The disparities between LLM-human agreement and human-level agreement ranged from $0.209$ to $0.352$ ($\kappa_{\text{Humans}} - \kappa_{\text{LLM}}$). This result is not surprising given the sparse instructions provided to the LLMs in the zero-shot prompt. In the second part of this study, we investigate whether and to what extent we can close this gap via prompt engineering.

\subsection{Improving LLM-Human Agreement}

Disagreement between LLMs and human raters generally stems from lack of alignment on thresholds separating the levels of a feature. Table~\ref{tab:proportion} shows the proportion of classifications made by the two human raters and o4-mini for each level of each feature. We can see that o4-mini is typically misaligned with the human raters in terms of how to separate the levels of a feature. For example, while human raters label a response as "fail[ing] to acknowledge or show sensitivity towards the legitimate concerns or feelings of one of the parties involved" (DISRES) only $5-6\%$ of the time, o4-mini classified $22.2\%$ of responses as such. Similarly, o4-mini classified $59.9\%$ of responses as having "Reasonable Justification", while classifying $0\%$ and $0.6\%$ of responses as displaying "No Justification" and "Clear and Compelling Justification", respectively. Human raters, meanwhile, provided more classifications at these extreme ends of the ordinal scale at the expense of labels in the middle of the scale. This result reflects an overall pattern we see across all non-binary features: o4-mini tended to provide more classifications in the middle of an ordinal scale than we observed with human raters.  We used these results to motivate our prompt engineering strategy and further delineate feature levels.

\begin{table*}[ht]
    \centering
    \begin{tabular}{c|c|ccc}
        \hline
        & & \multicolumn{3}{c}{\textbf{Proportion Selected}} \\
        \textbf{Key} & \textbf{Level} & \textbf{Human 1} & \textbf{Human 2} & \textbf{o4-mini} \\
        \hline
        \multirow{3}{*}{INT} & Limited Interpretation & 0.435 & 0.377 & 0.327 \\
                                 & Adequate Interpretation & 0.447 & 0.475 & 0.642 \\
                                 & Excellent Interpretation & 0.118 & 0.148 & 0.031 \\ \hline
        \multirow{2}{*}{LACKINF} & False & 0.944 & 0.889 & 0.864 \\
                                 & True & 0.056 & 0.111 & 0.136 \\ \hline
        \multirow{4}{*}{JUST} & No Justification & 0.062 & 0.056 & 0 \\
                                 & Superficial Justification & 0.358 & 0.333 & 0.395 \\
                                 & Reasonable Justification & 0.358 & 0.469 & 0.599 \\
                                 & Clear and Compelling Justification & 0.222 & 0.142 & 0.006 \\ \hline
        \multirow{2}{*}{VAGUE} & False & 0.790 & 0.883 & 0.568 \\
                                 & True & 0.210 & 0.117 & 0.432 \\ \hline
        \multirow{3}{*}{PERSP} & Considers one perspective & 0.302 & 0.407 & 0.149 \\
                                 & Briefly considers multiple perspectives & 0.407 & 0.549 & 0.758 \\
                                 & Thoughtfully considers multiple perspectives & 0.290 & 0.272 & 0.093 \\ \hline
        \multirow{2}{*}{DISRES} & False & 0.938 & 0.951 & 0.778 \\
                                 & True & 0.062 & 0.049 & 0.222 \\ \hline
        \multirow{2}{*}{CREAT} & False & 0.833 & 0.796 & 0.994 \\
                                 & True & 0.167 & 0.204 & 0.006 \\ \hline
    \end{tabular}
    \caption{Proportion of responses where each feature level was selected by human raters and o4-mini (with zero-shot prompt).}
    \label{tab:proportion}
\end{table*}

\begin{figure*}[t]
  \includegraphics[width=\linewidth]{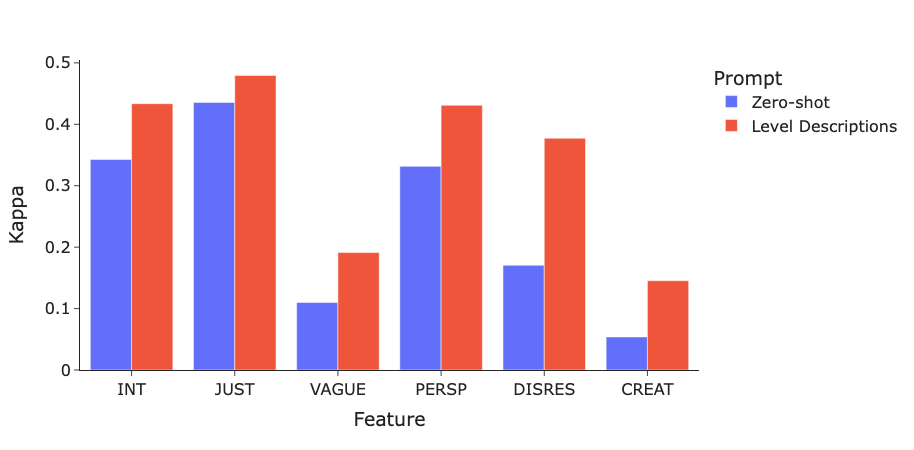}
  \caption {Average $\kappa$ with human raters using o4-mini with zero-shot prompting and prompting with additional details for each level of a feature. Human-LLM agreement improves when providing additional level details in the prompt.}
  \label{fig:compare_runs}
\end{figure*}

We focused on six features for prompt engineering, omitting LACKINF where o4-mini was already achieving close to human-level performance. Results are displayed in the last column of Table~\ref{tab:results} as well as Figure~\ref{fig:compare_runs}. We find that including additional details about the levels for a feature in the prompt effectively improves the LLM's agreement with humans. For most features we saw improvements of $0.08<\Delta\kappa<0.1$, but for DISRES we saw gains of $\Delta\kappa = 0.206$. With these prompts, o4-mini would've performed higher than all LLMs tested in the first experiment on all features except LACKINF (which we did not investigate improving) and CREAT, where o4-mini performs better, but is still outclassed by most other models.

\section{Conclusions and Future Work}

This study evaluated the feasibility of using LLMs to extract construct-relevant features from the Casper SJT. We found that reasoning models like OpenAI's o4-mini and Anthropic's Claude Sonnet 4 generally performed best at identifying these complex and nuanced constructs in responses, even with limited instructions. Additionally, we found that each LLM that we tested achieved the strongest performance on at least one feature. This result indicates that a future automated scoring solution using the same feature extraction method may be best served by using different LLMs for different features rather than forcing a single universal LLM. We could also consider using multiple LLMs for the same criteria and instituting a voting system resembling traditional machine learning ensemble methods to produce more accurate and reliable results~\cite{dietterich2000ensemble}. Overall, our results suggest a promising avenue for extracting construct-relevant features from SJTs and similar open-response assessments.

Prior to engaging in prompt engineering to improve performance, we had already reached close to human-level agreement in extracting one feature, whether a response "state[d] that they do not have enough information to make a decision." In this particular case, we hypothesize that features such as this one may be extractable by simpler models and methods such as keyword and semantic search.

For other features we fell well short of human-level agreement using zero-shot prompt classification, but we demonstrated that LLMs can be instructed to behave closer to expectations by giving further details about the levels for a feature. We found that providing these details had varying effects on our classification performance for different features, indicating that different features may be more susceptible to influence from this type of prompt engineering. Future work could explore other approaches to prompt engineering including few-shot prompting as well as fine-tuning to further improve performance.

We were also limited by small sample sizes in this study, owing to the effort and expertise required to annotate datasets such as these based on the features we explored. Future work will extend this study to explore a larger dataset from the Casper SJT as well as additional features. We plan to investigate the use of these features in an eventual automated scoring system for the Casper SJT. Such work would have important consequences, potentially extending the scalability and standardization of open-response assessments of personal and professional skills.

An automated scoring system based on the approach demonstrated here would also open avenues for future work in formative assessments by providing real time evaluation and feedback to respondents. We used a system prompt in this study that returned both a "decision" and "reasoning". We did not inspect the "reasoning"s in this study, but future work could use these "reasoning" fields to generate personalized and direct feedback for respondents. This method of extracting features from text could also be extended beyond assessments to other pieces of written text such as personal essays and reference letters.

\section*{Acknowledgments}

We would like to acknowledge Jillian Derby, Alexander MacIntosh, Josh Moskowitz, Gill Sitarenios, Susha Suresh, and Yiyu Xie for thoughtful discussions on the ideas presented in this manuscript.

\bibliography{references}

\end{document}